\documentclass{article} 
\usepackage[preprint]{colm2025_conference}

\usepackage{microtype}
\usepackage{hyperref}
\usepackage{url}
\usepackage{booktabs}
\usepackage{tabularx}
\usepackage{graphicx}
\usepackage{subcaption}

\usepackage{lineno}

\usepackage[ruled]{algorithm2e}
\usepackage{bm}

\definecolor{darkblue}{rgb}{0, 0, 0.5}
\hypersetup{colorlinks=true, citecolor=darkblue, linkcolor=darkblue, urlcolor=darkblue}

\title{Pre-Training LLMs on a budget: A comparison of three optimizers}

\author{Joel Schlotthauer, Christian Kroos, \\\textbf{Chris Hinze, Viktor Hangya, Luzian Hahn \& Fabian Küch} \\
Fraunhofer Institute for Integrated Circuits IIS, Erlangen, Germany\\
name.surname@iis.fraunhofer.de\\
}

\begin{document}

\ifcolmsubmission
\linenumbers
\fi

\maketitle

\begin{abstract}
Optimizers play a decisive role in reducing pre-training times for LLMs and achieving better-performing models. In this study, we compare three major variants: the de-facto standard \textit{AdamW}, the simpler \textit{Lion}, developed through an evolutionary search, and the second-order optimizer \textit{Sophia}. For better generalization, we train with two different base architectures and use a single- and a multiple-epoch approach while keeping the number of tokens constant. Using the Maximal Update Parametrization and smaller proxy models, we tune relevant hyperparameters separately for each combination of base architecture and optimizer. We found that while the results from all three optimizers were in approximately the same range, Sophia exhibited the lowest training and validation loss, Lion was fastest in terms of training GPU hours but AdamW led to the best downstream evaluation results.

\end{abstract}

\section{Introduction}
\label{introduction}

Pre-training Large Language Models (LLMs) requires significant computing resources. Even for relatively small models with around 3 billion parameters, thousands of GPU hours are necessary to achieve meaningful language understanding and generation performance. Not surprisingly, methods that help to reduce the amount of training time and effort have gained traction in LLM pre-training in general and are inevitable if the compute budget is severely limited. Considering the usual training dynamics, the learning rate emerges as the most decisive hyperparameter. A fixed learning rate over the entire course of the training, however, cannot escape the disadvantageous trade-off relationship between slowing down the convergence progress in the first halve of the training and missing the (local) optimum towards the end of the training: either the weight update steps are too small or too big. Therefore, dynamic learning rate adjustments were introduced in the form of schedulers and dynamic adaptations of the impact of the learning rate in the form of the optimizers \citep{wen+2024}, which optimize the gradient descend itself \citep{ruder+2017}. The de-facto standard for the latter is currently \textit{AdamW} \citep{loshchilov+2017}, a modification of the also popular \textit{Adam} \citep{kinga+2015} that uses momentum to adjust the learning rate (see Algorithm \ref{alg:adamw}).

Despite AdamW’s dominance, many variants \citep[e.g., ][]{zhang+2018, reyad+2023, zhang+2024} and alternatives have been proposed \citep[e.g., ][]{shazeer+2018, bernstein+2018, zhuang+2020, kwon+2021, wright+2021, xie+2024}. This study compares two of them, \textit{Lion} \citep{chen+2023} and \textit{Sophia} \citep{liu+2024}, with AdamW, in the context of small models and considering a limited compute budget. \textit{Lion} was selected for its simplicity and unusual origin as the result of an evolutionary search, that is, it was not designed by a human engineer (see Algorithm \ref{alg:lion}). \textit{Sophia} was chosen because it employs second-order criteria while maintaining a computationally lean approach (see Algorithm \ref{alg:sophia}). We unified the display of the three algorithms in order to make it easier to see their commonalities and differences.  

\newcommand\mycommfont[1]{\footnotesize\sffamily\textcolor{gray}{#1}}
\SetCommentSty{mycommfont}
\SetKwComment{Comment}{$\triangleright$ }{ }
\SetKwInput{KwParameter}{Hyperparameters:} 
\SetKwInput{KwNetwork}{Network function:} 
\SetKwInput{KwEstimator}{Estimators:} 
\SetKwInput{KwInitialize}{Initialize:} 
\SetKwInput{KwReturn}{Return:} 

\begin{algorithm}
    \DontPrintSemicolon
    \caption{AdamW}\label{alg:adamw}
    \small
    \KwParameter{first moment factor $\beta_1$, second moment factor $\beta_2$, learning rate $\eta$, small constant (to avoid division by $0$) $\epsilon$, weight decay factor $\lambda$}
    \KwNetwork{$f$}
    \KwInitialize{}
        \hspace{10pt} step $t \gets 0$\;
        \hspace{10pt} parameter vector $\bm{\theta_{t = 0}} \in \mathbb{R}^n$\; 
        \hspace{10pt} first moment vector $\bm{m}_{t = 0} \gets \bm{0}$\;
        \hspace{10pt} second moment vector $\bm{v}_{t = 0} \gets \bm{0}$\;    
    \While{$\bm{\theta_t}$ not converged}{
        $t \gets t + 1$\;
        $\bm{g}_t \gets \nabla f_t(\bm{\theta}_{t-1})$ \Comment*[r]{obtain gradient}
        $\bm{m}_t \gets \beta_1 \bm{m}_{t-1} + (1 - \beta_1) \, \bm{g}_t$ \Comment*{update first moment using gradient}
        $\bm{v}_t \gets \beta_2 \bm{v}_{t-1} + (1 - \beta_2) \, \bm{g}_t^2$ \Comment*[r]{update second moment using square of gradient}
        $\bm{\hat{m}}_t \gets \bm{m}_t / (1 - \beta_1^t)$ \Comment*[r]{first moment bias correction}
        $\bm{\hat{v}}_t \gets \bm{v}_t / (1 - \beta_2^t)$ \Comment*[r]{second moment bias correction}
        $\bm{\theta}_t \gets \bm{\theta}_{t-1} - \eta_t(\alpha \,\bm{\hat{m}}_t   
        /(\sqrt{\bm{\hat{v}}_t + \epsilon}) + \lambda \, \bm{\theta}_{t-1})$ \Comment*[r]{update model parameters with weight decay}  
    }
    \KwReturn{$\bm{\theta_t}$}
\end{algorithm}


Our aim was to investigate for auto-regressive decoder-only language models, whether any of the three optimizers would outperform the others, tracked separately for training and validation loss, as well as wall clock time and downstream performance on several common LLM benchmarks. We were specifically interested in smaller models of the approximate size of 3 billion parameters trained with 60 billion tokens, following the Chinchilla scaling laws \citep{hoffmann+2022}, under the assumption of a limited computational training budget. Furthermore, we were interested whether the early indicators, the two loss measures, would predict downstream performance, the most important measure for model development. To ensure greater generality of the results, we additionally applied a multiple-epoch approach (while keeping the number of tokens overall the same as in the single-epoch approach) and employed two different main-stream model architectures. Importantly, we optimized all hyperparameters for each optimizer and architecture separately in order to enable an adequate comparison via the best possible outcome. Since a grid search even with only few hyperparameters and a model with approximately 3 billion parameters is not feasible given the usual limits in the compute budget, we employed the Maximal Update Parametrization \citep[$\mu$P, ][]{yang+2021} to find optimal values using small proxy models (50 million parameters) and then transfer these values unchanged to the larger target model (2.7 billion parameters). Based on theoretical findings, $\mu$P involves a minor re-parametrization of the neural network representation that guarantees invariance of the optimum for the most important hyperparameters to size changes of the model. 
We implemented $\mu$P for Lion and Sophia and established in a series of auxiliary experiments that it works with these two optimizers as reliably as with AdamW. 

Our contributions are:
\begin{itemize}
    \item An in-depth comparison of three major optimizers (AdamW, Lion, Sophia) covering the entire evaluation line from training loss to application benchmarks;
    \item A partial generalisation of the findings by confirming them with multiple-epoch training and an additional architecture type;
    \item A consistent methodological approach for fine-tuning hyperparameters for each optimizer or optimizer-architecture combination enabling an unbiased comparison; 
    \item Empirical evidence for the ability of effective hyperparameter transfer under the Maximal Update Parametrization with Lion and Sophia.
\end{itemize}

\section{Related work}
\label{related_work}
When a new optimizer is proposed, good scientific and engineering practices require a comparison of the new optimizer with established alternatives. Accordingly, the authors of the respective papers of Lion and Sophia compared their optimizer with AdamW and in the case of Sophia also with Lion. When contrasting Lion with AdamW on three model sizes (1.1, 2.1, 7.5 billion parameters) and 24 NLP tasks (both NLU and NLG) with one-shot evaluation, \citep{chen+2023} found that the models trained with Lion outperformed those trained with AdamW. This was not confirmed by \citep{liu+2024} who trained a small model (355M parameters) using all three optimizers and evaluated them with a few-shot evaluation on SuperGLUE. Overall AdamW came out slightly superior to Lion but inferior to the author's own Sophia. For two larger models they compared only AdanW with Sophia and Sophia resulted in a better performance in most tasks. 
However, common to these studies (and similar ones proposing other optimizers) is a familiarity advantage of the authors regarding their own creation which might explain differences in the results. It can be speculated that based on a deeper theoretical understanding and the extensive expertise gained during the development, tuning the hyperparameters of the novel optimizer and other model and training hyperparameters is more successful than with the optimizers from the literature with which they are compared. 

\begin{algorithm}
    \DontPrintSemicolon
    \caption{Lion}\label{alg:lion}
    \small
    \KwParameter{first moment factors $\beta_1$ \& $\beta_2$, learning rate $\eta$, weight decay factor $\lambda$}
    \KwNetwork{$f$}
    \KwInitialize{}
        \hspace{10pt} step $t \gets 0$\;
        \hspace{10pt} parameter vector $\bm{\theta_{t = 0}} \in \mathbb{R}^n$\; 
        \hspace{10pt} first moment vector $\bm{m}_{t = 0} \gets \bm{0}$\;
    \While{$\bm{\theta_t}$ not converged}{
        $t \gets t + 1$\;
        $\bm{g}_t \gets \nabla f_t(\bm{\theta}_{t-1})$ \Comment*[r]{obtain gradient}
        $\bm{c}_t \gets \beta_1 \bm{m}_{t-1} + (1 - \beta_1) \, \bm{g}_t$ \Comment*{update first moment using gradient}
        $\bm{\theta}_t \gets \bm{\theta}_{t-1} - \eta_t(\alpha \, $sign$(\bm{c}_t) + \lambda \, \bm{\theta}_{t-1})$ \Comment*[r]{update model parameters with weight decay}     
        $\bm{m}_t \gets \bm{m}_{t -1} + (1 - \beta_2) \, \bm{g}_t$ \Comment*{update first moment again using gradient}
    }
    \KwReturn{$\bm{\theta_t}$}
\end{algorithm}

Looking at independent comparative studies, to the best of our knowledge, only two have been published, namely \citet{kaddour+2023} and \citet{zhao+2025}. 

\citet{kaddour+2023} evaluated AdamW, Lion and Sophia tracking training and validation loss and measuring the performance of three downstream tasks. All three optimizers were found to perform approximately equally well with the exception of a few tasks where AdamW excelled relative to the poor performance of Lion and Sophia. However, the models they used were the encoder-only BERT model trained with a masked language modeling (MLM) objective and the encoder-decoder T5 model trained with a span-corrupting MLM objective. Both models were very small compared with recent LLMs, consisting of 120M parameters (BERT) and 220M parameters (T5), respectively. It is therefore unverified whether their results hold for larger decoder-only models trained with auto-regressive next-token prediction. In addition, the authors fine-tuned the trained models for the downstream tasks (GLUE and SuperGLUE for BERT, ROUGE-L for T5) potentially confounding the impact of the different optimizers in pre-training.

\begin{algorithm}
    \DontPrintSemicolon
    \caption{Sophia}\label{alg:sophia}
    \small
    \KwParameter{first order moment factor $\beta_1$, second order moment factor $\beta_2$, step interval between Hessian estimates $k$, learning rate $\eta$, weight decay factor $ \lambda$, factor controlling fraction of clipped entries $\gamma$, small constant (to avoid division by $0$) $\epsilon$}
    \KwEstimator{Est $\in \{$Hutchinson, Gauss-Newton-Bartlett$\}$}
    \KwNetwork{$f$}
    \KwInitialize{}
        \hspace{10pt} step $t \gets 0$\;
        \hspace{10pt} parameter vector $\bm{\theta_{t = 0}} \in \mathbb{R}^n$\; 
        \hspace{10pt} first moment vector $\bm{m}_{t = 0} \gets \bm{0}$\;
        \hspace{10pt} second order vector $\bm{h}_{k} \gets \bm{0}$\;
    \While{$\bm{\theta_t}$ not converged}{
        $t \gets t + 1$\;
        $\bm{g}_t \gets \nabla f_t(\bm{\theta}_{t-1})$ \Comment*[r]{obtain gradient} 
        $\bm{m}_t \gets \beta_1 \bm{m}_{t-1} + (1 - \beta_1) \, \bm{g}_t$ \Comment*{update first order moment using gradient}
        \eIf{$t$ mod $k$}{
            $\bm{\hat{h}}_t \gets $ Est$(\bm{\theta}_{t - 1})$ \Comment*[r]{obtain estimate of diagonal Hessian}  
            $\bm{h}_t \gets \beta_2 \bm{h}_{t-k} + (1 - \beta_2) \, \bm{\hat{h}}_t$ \Comment*{update second order moment using estimate}
        }{
            $\bm{h}_t \gets \bm{h}_{t - 1}$ \Comment*{retain updated estimate from last step}
        }
        $\bm{\theta}_t \gets \bm{\theta}_{t-1} - \eta_t  \lambda \, \bm{\theta}_{t-1}$ \Comment*{apply weight decay}  
        $\bm{\theta}_t \gets \bm{\theta}_{t} - \eta_t \cdot $ 
        clip$(\bm{m}_t / $max$ \{ \gamma \, \bm{h}_t, \; \epsilon \}, \;1 )$ \Comment*[r]{update model parameters}  
    }
    \KwReturn{$\bm{\theta_t}$}
\end{algorithm}

\citet{zhao+2025} investigated the performance of SGD, Adafactor, Adam, Lion, and Sophia for auto-regressive models of different smaller model sizes (150M, 300M, 600M, and 1.2B parameters) and a few hyperparameter variations. They found little differences in the performance of the optimizers and their stability across different learning rates. They did not use $\mu$P and did not optimize the hyperparameters for each optimizer but instead explored the impact of several hyperparameters with one-dimensional sweeps. However, they did not conduct any downstream evaluations. For the smaller models those might have been problematic in the first place and even for their biggest model difficult: At least some of the common benchmarks require a certain level of capability which to date only larger models can provide. Thus, the authors relied on validation loss as the single evaluation measure, which might not be sufficient to gauge the impact of an optimizer when it comes to real-world applications.

\section{Method}
\label{method}
\subsection{Model architecture}
Our choice for the two types of decoder-only architecture in the study was inspired by contrasting the GPT-2 family \citep{radford+2019, brown+2020} and the LLaMA family \citep{touvron+2023} of models. GPT-2 uses two linear layers in each of its Multi-Layer-Perceptrons (MLP) while LLaMA uses three, although the overall number of neurons can be kept the same in both architectures. GPT-2 uses layer normalization whereas LLaMA is characterized by RMS normalization. In addition, GPT-2 incorporates bias vectors as input to all layers, while LLaMA eschews them, and GPT-2 employs GELU as activation function, while LLaMA uses SwiGLU. In the following we will refer to the two architectures as GPT-2 and a LLaMA architecture. We set the target model size to approximately 3 billion parameters and the proxy model size for hyperparameter training to approximately 50 million parameters. Table \ref{tab:model_specs} shows the specifications for the four models. For all models, the context size was set to 2048 and RoPE \citep{lee+2021} was used for positional encoding. 

\subsection{Dataset and tokenization}

We use the publicly available SlimPajama dataset \citep{soboleva+2023} for all experiments. The full dataset comprises roughly 627B tokens, from which we generated a random shuffle and subsequently selected a contiguous slice of 60B tokens for training. For multi-epoch runs, we further extracted a subset of 12B tokens from this 60B-token slice. All models, including LLaMA-based architectures, were trained using the GPT-2 tokenizer to ensure consistency and isolate the effects of optimizer choice, architecture differences, and hyperparameter tuning.


\subsection{Hardware and software setup}
All model variants are trained using 2 cluster nodes equipped with 8 NVIDIA A100~80GB GPUs each (16 GPUs total) per configuration. We build upon a custom fork of the popular LLM toolbox \textsc{LitGPT} \citep{litgpt+2023}, that enables us flexibility in model definition and training scripts. The primary libraries used include: PyTorch~2.3.2 for core neural-network operations; CUDA~12.3 for GPU acceleration. We train in BF16 mixed-precision and utilize Fully Sharded Data Parallel (FSDP) for parallelization. We use the Flash Attention 2 algorithm built in Pytorch 2.3.2 and utilize torch compile to further speed up our training by 30\%. We use the standard Pytorch implementation of AdamW, the implementation of Sophia by their authors \footnote{https://github.com/Liuhong99/Sophia} and the \textsc{lion-pytorch} implementation by lucid-rains \footnote{https://github.com/lucidrains/lion-pytorch}.

\begin{table}[t]
    \begin{center}
    \begin{tabular}{lllll}
        \toprule
            &  \bf GPT-2 2.7B  & \bf LLaMA 2.7B  & \bf GPT-2 50M  & \bf LLaMA 50M  \\ 
        \midrule
        Layers             & 32      & 32       & 32       & 32     \\ 
        Number of heads    & 32      & 32       & 4        & 4      \\ 
        Head size          & 80      & 80       & 64       & 64     \\ 
        Embedding size     & 2560    & 2560     & 256      & 256         \\ 
        Normalization class    & LayerNorm  & RMSNorm  & LayerNorm  & RMSNorm  \\ 
        Bias    & True         & False         & True         &  False        \\ 
        MLP class    & two-layer  &   three-layer  &  two-layer   & three-layer  \\ 
        Activation function    & GELU  &   SwiGLU  &  GELU   & SwiGLU  \\ 
    \bottomrule
    \end{tabular}
    \end{center}
    \caption{Model specification for target and proxy models.}
    \label{tab:model_specs}
\end{table}

\subsection{Hyperparameter tuning}
\label{sec:hyperparam-tuning}

We use the Maximal Update Parametrization \citep[$\mu$P, ][]{yang+2021} to optimize hyperparameters on very small proxy models and subsequently transfer these hyperparameters directly to larger target models—ensuring that the optimal hyperparameters identified at smaller scale remain optimal at larger scales ($\mu$Transfer). While $\mu$P rules have been established and validated for AdamW \footnote{https://github.com/microsoft/mup}, corresponding setups and empirical validations for Lion and Sophia optimizers remained inconclusive for Lion \citep{lingle+2025} and were previously unaddressed for Sophia.
In $\mu$-parametrization, a scaling scheme is applied across three aspects: (i) the initialization variance (for input, output, or hidden weights), (ii) the learning rate (which may depend on the optimizer variant), and (iii) any activation multipliers (also see Table \ref{tab:mup-scaling-rules}). Although the scale factor is derived from the parameter’s input- and output dimensions, in practice it is simply derived from the width ratio between the tuned proxy model and the larger target model.
Accordingly, we implemented $\mu$P for Lion and Sophia and showed in a series of auxiliary experiments that $\mu$Transfer is valid when using these optimizers (see next Section, \ref{sec:mup_with_lion_and_sophia}).

\paragraph{Method}
We conducted a grid search using proxy models with 50 million parameters, separately optimizing hyperparameters for each of the combinations defined by optimizer (AdamW, Lion, Sophia) and architecture (GPT-2, LLaMA). Proxy models were trained on an 1B-token subset from the same shuffled SlimPajama dataset later employed for full-scale training, using a total batch size of 1M tokens. Optimal hyperparameters were identified by selecting the configuration that minimized final training loss from a grid of 365 options per system. These selected hyperparameters were directly transferred to our larger 2.7B-parameter target models trained with a batch size of 4M tokens.

\paragraph{Tuned hyperparameters choices}

We chose the learning rate as the most critical hyperparameter to tune, followed by a multiplier of the activations of the final layer (for additional rationale, see Appendix \ref{appendix:hyperparameter-selection}). For Sophia, we also identified $\rho$, controlling the influence of second-order information, as important, prompting us to slightly reduce the resolution of our learning-rate search for Sophia to maintain overall tuning consistency.
We use optimizer-specific hyperparameters ($\epsilon$, $\beta_1$, $\beta_2$) chosen based on recommendations from the original publications (further details in Table \ref{tab:hyperparams_summary}). The initialization variance was fixed at 0.073 based on consistent evidence from prior experiments with models of similar architecture and scale, aligning with findings in the literature \citep{dey2023btlm}.


\subsection{Validation of $\mu$P with Lion and Sophia}
\label{sec:mup_with_lion_and_sophia}

To our knowledge, successful hyperparameter transfer ($\mu$Transfer) under $\mu$P has not previously been validated for Lion and Sophia optimizers. We thus empirically assessed $\mu$Transfer across model widths ranging from 128 to 2048, exploring a wide range of learning rates to identify consistent optima.The $\mu$P framework defines distinct parameterization regimes primarily influencing scaling rules for input, output, and hidden-layer weights (Table~\ref{tab:mup-scaling-rules}). While established scaling rules exist for AdamW, this study extends the empirical validation of $\mu$Transfer to Lion and Sophia, optimizers structurally related to SignSGD and AdamW respectively.

\paragraph{Implementation}  
We streamlined the original $\mu$P implementation for compatibility with Fully Sharded Data Parallelism (FSDP). Instead of continuously referencing precomputed scaling information (infshapes) during training, we leveraged infshapes solely during model initialization by dynamically overriding module initialization routines (\texttt{reset\_parameters()}). Scaling factors were subsequently stored in model configurations, enabling efficient runtime learning-rate scaling without further infshape references.

\paragraph{Method}  
We evaluated optimal learning rates for Lion and Sophia at widths of 128, 256, 512, 1024, and 2048, training each model configuration for 800M tokens of the SlimPajama dataset. Optimal learning rates were identified based on minimizing average training loss over the final 5\% of training steps.

\paragraph{Results and Discussion}  
Our empirical results indicate successful hyperparameter transfer under $\mu$P for both Lion and Sophia, as the identified optimal learning rates remained consistent across model widths within our experimental setup. For Sophia, optimal $\mu$Transfer occurred under the AdamW scaling regime, consistent with structural similarities between the optimizers (further elaborated in Appendix~\ref{appendix:mup-validation}). Interestingly, Lion also exhibited optimal transfer performance under AdamW scaling, despite greater conceptual similarity to SignSGD. However, we observed significant sensitivity around the optimum learning rate, particularly for Lion, indicating that careful hyperparameter tuning remains essential to achieve effective $\mu$Transfer. Details on optimizer-specific scaling regimes and sensitivity analyses are provided in Appendix~\ref{appendix:mup-validation}, including a summary of scaling rules adapted from \citet{yang+2021}.

\subsection{Model training}
\label{model_training}

We trained models using three optimizers (AdamW, Lion, Sophia) across two architectures (GPT, LLaMA), exploring two distinct data regimes: unique data (single epoch over 60B tokens) and repeated data (five epochs over 12B tokens, totaling 60B tokens). Due to computational constraints, only GPT models were trained on the unique-data regime to establish baseline performance. Both GPT and LLaMA models were subsequently trained under the repeated-data regime, allowing analysis of generalization effects across different data conditions and architectures.

All training runs employed a batch size of 4M tokens using gradient accumulation, distributed via hybrid FSDP sharding across 16 A100 GPUs. The learning rate schedule featured a linear warmup phase over the initial 750M tokens, followed by linear decay to 10\% of the peak value. Hyperparameters determined via preliminary tuning on smaller models transferred robustly to these main experiments, resulting in stable training dynamics.

\subsection{Evaluation}
\label{evaluation}

We evaluated optimizer performance using both training-related metrics and downstream evaluation scores. For training, we tracked final training and validation losses (negative log-likelihood), using a fixed SlimPajama subset for validation across all runs. To reduce noise, losses were averaged over the final 5\% of training, where loss curves had converged. We also computed the area under the training loss curve (AULC) to assess convergence speed, and recorded total training times. For downstream evaluation, we benchmarked models trained for 60B tokens on standard tasks using the \emph{lm-evaluation-harness} framework~\citep{eval-harness}. As small models trained on fewer than 100B tokens have limited capabilities, we focused on accuracy for ARC-Easy, ARC-Challenge~\citep{Clark2018ThinkYH}, Hellaswag~\citep{zellers2019hellaswag}, and MMLU~\citep{hendryckstest2021} in a zero-shot setup. Following~\citet{gu2024olmes}, we used the cloze formulation for MMLU—ranking answer options by LM probability—rather than the multiple-choice variant, which small models typically struggle with.

\section{Results}
\label{results}

\subsection{Training dynamics}

\begin{figure}[t]
\begin{center}
\includegraphics[width=0.8\linewidth]{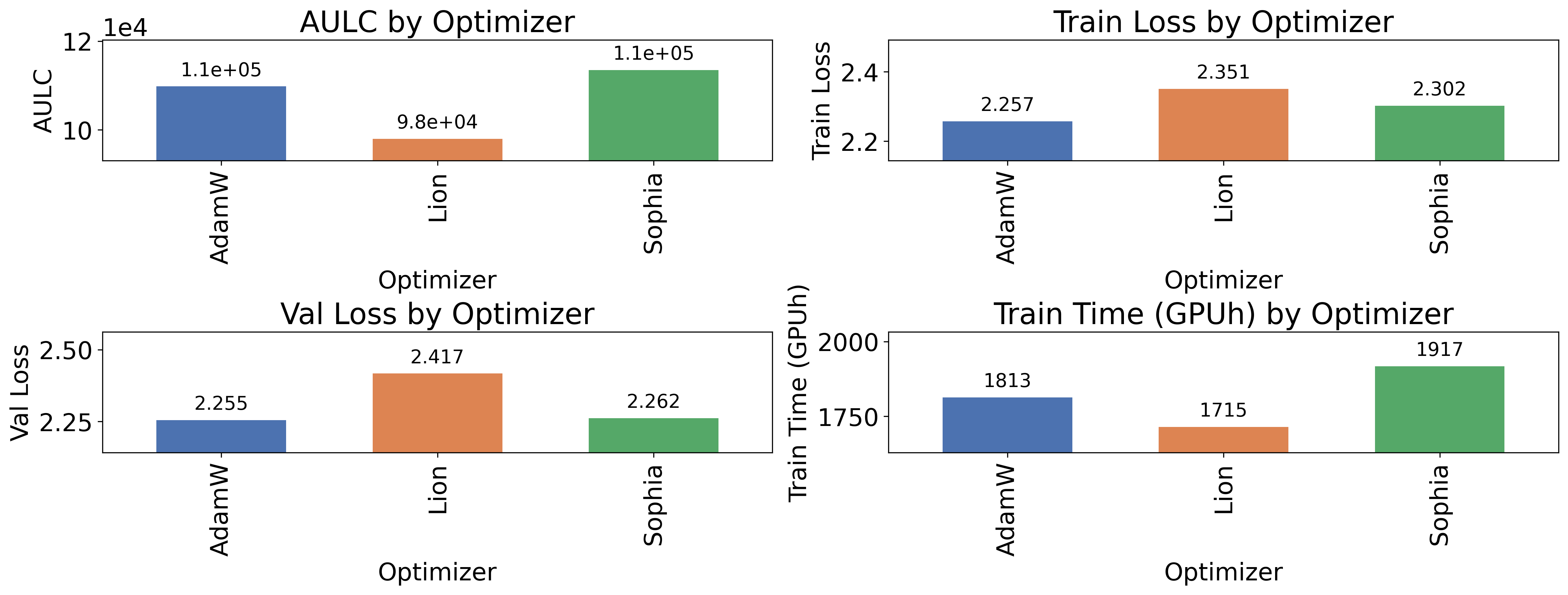}  
\end{center}
\caption{Training dynamics of optimizers on GPT models trained with 60B unique tokens.}
\label{fig:training_dynamics_results_overview_60b_unique}
\end{figure}
\begin{figure}[t]
\begin{center}
\includegraphics[width=0.8\linewidth]{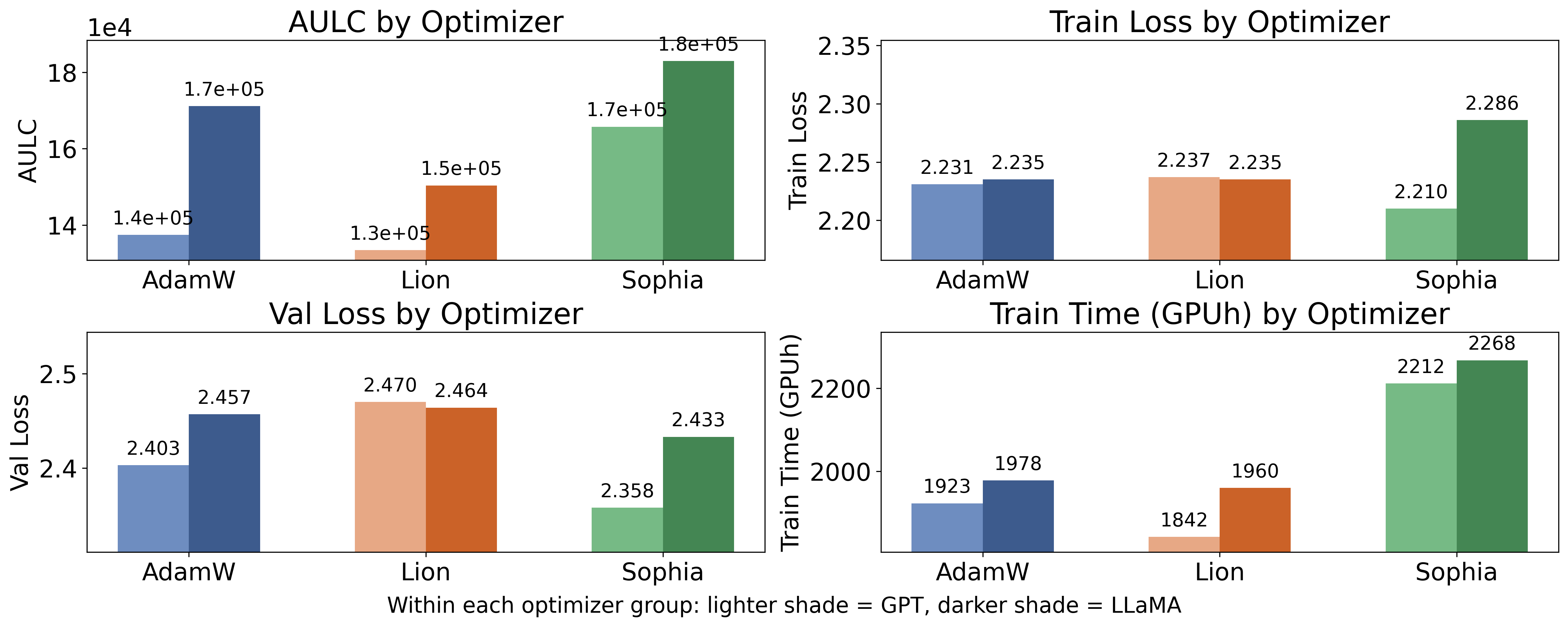}  
\end{center}
\caption{Training dynamics of optimizers across GPT and LLaMA architectures trained over five epochs (totaling 60B tokens). \textit{Note:} Sophia’s training times were inflated due to temporary file system slowdowns. For more representative Sophia timings, see Figure~\ref{fig:training_dynamics_results_overview_60b_unique}.}

\label{fig:training_dynamics_results_overview}
\end{figure}

In the single-epoch regime (Figure~\ref{fig:training_dynamics_results_overview_60b_unique}), Lion exhibited the fastest initial convergence, achieving the lowest AULC and shortest training duration, but fell behind in later stages. AdamW and Sophia showed similar AULC values; however, AdamW achieved the lowest final training and validation losses, indicating superior generalization. In the multi-epoch regime (Figure~\ref{fig:training_dynamics_results_overview}), Lion again yielded the lowest AULC across both architectures, though performance varied between architectures, favoring LLaMA. Sophia benefited from repeated epochs, especially with GPT architectures, achieving the lowest overall losses, albeit with slightly increased (~6\%) computational overhead to AdamW to second-order computations. AdamW maintained consistently balanced performance. Generally, GPT architectures outperformed LLaMA, except with Lion, where the trend reversed.

\subsection{Downstream performance}

\begin{figure}[t]
\begin{center}
\includegraphics[width=0.8\linewidth]{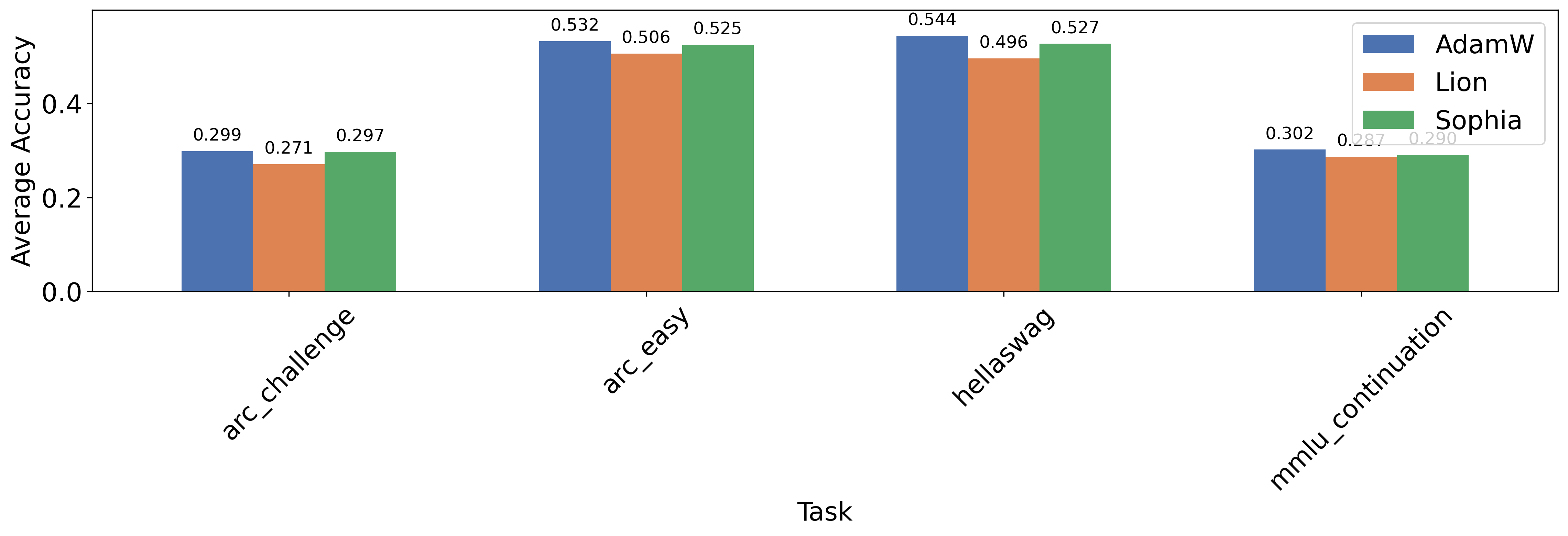}  
\end{center}
\caption{Downstream performance of AdamW, Lion and Sophia with the GPT architecture trained on 60B unique tokens on four common tasks.}
\label{fig:downstream_eval_optimizers_60b_unique}
\end{figure}

\begin{figure}[t]
\begin{center}
\includegraphics[width=0.8\linewidth]{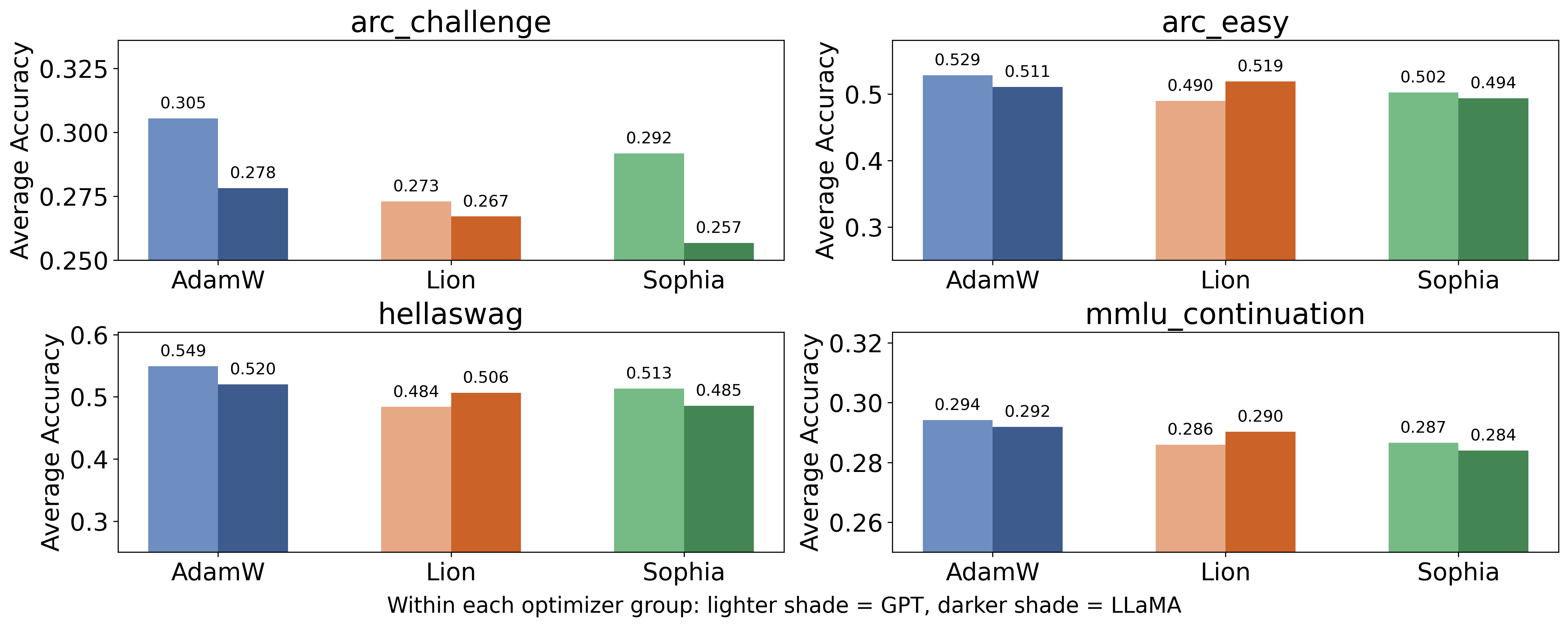}  
\end{center}
\caption{Downstream task performance of the three optimizers with both architectures, GPT and LLaMA, trained on five epochs of 12B tokens.}
\label{fig:downstream_eval_results_overview}
\end{figure}

AdamW consistently achieves the highest downstream accuracy across benchmarks, clearly outperforming both Lion and Sophia (Figures~\ref{fig:downstream_eval_optimizers_60b_unique},~\ref{fig:downstream_eval_results_overview}). GPT architectures typically yield higher downstream accuracy than LLaMA counterparts, except for Lion, where LLaMA performs slightly better on average (Figure \ref{fig:downstream_eval_results_overview}). 

\begin{figure}[t]
\begin{center}
\includegraphics[width=0.8\linewidth]{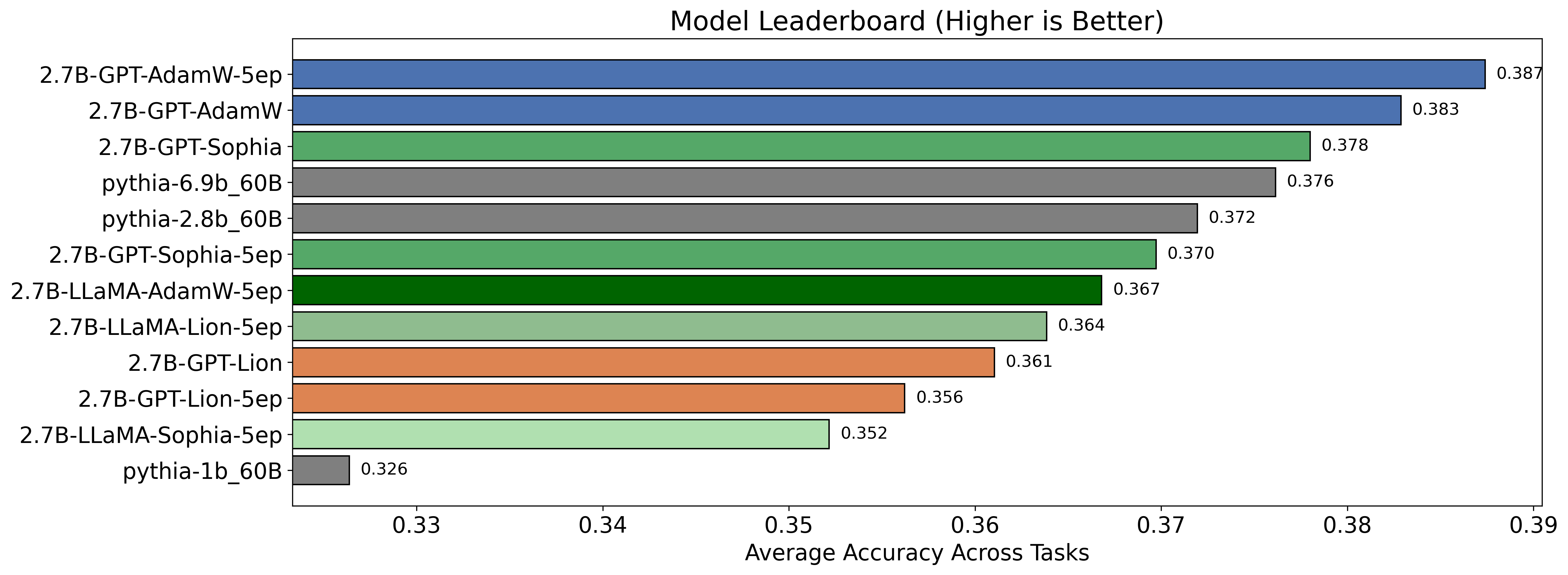}  
\end{center}
\caption{Leaderboard comparison of downstream performance (averaged across tasks). Models marked '5ep' were trained for 5 epochs on 12B tokens (total 60B tokens). Pythia checkpoints (60B tokens) included for comparison.}
\label{fig:downstream-leaderboard}
\end{figure}

Models trained on unique tokens (60B) generally outperform or match the multi-epoch variants (Figure \ref{fig:downstream-leaderboard}); with the exception of AdamW, achieving marginally higher downstream performance in the multi-epoch regime. 

Our best-performing models surpasses publicly available Pythia \citep{biderman2023pythiasuiteanalyzinglarge} checkpoints, outperforming both similarly sized (2.8B) and larger (6.9B) Pythia models (Figure \ref{fig:downstream-leaderboard}). This highlights the effectiveness of our training setups, though direct comparisons are nuanced due to differences in dataset composition.

\section{Discussion}
\label{discussion}

In this work, we systematically compared three optimizers—AdamW, Lion, and Sophia—in the context of pre-training small-scale autoregressive language models (approximately 3 billion parameters) under limited computational resources. Prior research provided limited guidance on optimizer selection specifically tailored to smaller models and constrained compute budgets, particularly regarding downstream task performance and generalization across data regimes and architectures. Additionally, the viability of hyperparameter transfer using the Maximal Update Parametrization ($\mu$P) was previously established primarily for AdamW but remained unexplored or inconclusive for Lion and Sophia.

Our empirical results address these gaps by highlighting distinct trade-offs among the tested optimizers. Lion demonstrated rapid convergence and computational efficiency, ideal for resource-constrained scenarios or rapid experimentation. In contrast, Sophia, though slightly more computationally demanding, achieved superior final validation losses, particularly under repeated data epoch conditions with GPT architectures. AdamW, however, consistently provided a robust balance by excelling notably in downstream task performance benchmarks, reinforcing its practical value as a standard choice.

Repeating epochs can enhance downstream performance, particularly when employing AdamW, a finding consistent with evidence suggesting multi-epoch training can even benefit performance under data constraints \citep{muennighoff2023scaling}. Architecturally, GPT-based models responded effectively to optimization strategies with Sophia and AdamW; however, for Lion, the LLaMA architecture achieved better results. Notably, optimal learning rates for LLaMA models required approximately twice the values identified for GPT counterparts, underscoring the importance of architecture-specific tuning.

Furthermore, we empirically validated hyperparameter transferability using $\mu$P for Lion and Sophia via the AdamW-based $\mu$P scaling rules, extending previously established results.

Practitioners aiming to pre-train small-scale models within limited budgets should carefully consider optimization goals, computational constraints, chosen architecture, and data regime. Lion is particularly attractive for computational efficiency and rapid iteration; Sophia excels in validation loss reduction during multi-epoch training; AdamW remains optimal for maximizing downstream task performance.

By systematically exploring these optimizers, data regimes, and architectures, we provide practical recommendations and a foundation for informed optimizer selection under constrained computational scenarios.

\section{Limitations}
\label{limitations}
Though being sizable with 2.7 billion parameters, our models are still per study design on the smaller side when it comes to being suitable as general purpose LLMs. It is unlikely that the findings would change for models twice or three times the size but it cannot be guaranteed for even larger models, e.g., in the 70 billion parameter range. As alluded to in Section \ref{introduction}, even models of around 'only' 3 billion parameters require a substantial amount of GPU hours to be trained. Our compute budget did not allow to repeat the experiments several times with different random seeds and/or repeated random sampling of the batches. Thus, we could not employ statistical inference tests and provide confidence intervals for the observed differences. However, preliminary results of an ongoing study at our department with a focus on random variation in language models points toward low variances at least regarding training loss if the hyperparameters and specifically learning rate were optimized (using $\mu$P).


\section*{Acknowledgments}
This work was funded by the German Federal Ministry for Economic Affairs and Climate Action
(BMWK) through the project OpenGPT-X (project no. 68GX21007D)

The authors gratefully acknowledge the scientific support and HPC resources provided by the Erlangen National High Performance Computing Center (NHR@FAU) of the Friedrich-Alexander-Universität Erlangen-Nürnberg (FAU) under the NHR project "ALLMT: Acceleration of Large Language Model Training". NHR funding is provided by federal and Bavarian state authorities. NHR@FAU hardware is partially funded by the German Research Foundation (DFG) – 440719683.

\bibliography{colm2025_conference}
\bibliographystyle{colm2025_conference}

\appendix
\section{Appendix}
\subsection{$\mu$P discussion and validation for Lion and Sophia}
\label{appendix:mup-validation}
We validated hyperparameter transferability across optimizers using $\mu$P. For AdamW, optimal transfer under $\mu$P involves scaling the learning rate inversely proportional to the model's width multiplier for hidden weights while maintaining constant weight decay. Sophia closely mirrors AdamW structurally; therefore, identical scaling rules effectively apply despite Sophia's diagonal Hessian approximation every $k$ steps \citep{ishikawa2024parameterizationsecondorderoptimizationeffective}.

For Lion, despite its similarity to SignSGD, AdamW-style learning-rate scaling achieves near-optimal transfer. Further improvements emerged by scaling weight decay proportionally to the width multiplier.

Notably, our results showed consistent hyperparameter scaling behaviors between GPT-2 and LLaMA architectures, aligning with $\mu$P's robustness to architectural variations such as bias inclusion or normalization method (LayerNorm vs. RMSNorm).

We note that while both Lion and Sophia follow AdamW-style scaling rules, hyperparameter transfer across these optimizers is less robust over a wide range of learning rates than for AdamW. Robust transfer is achieved only when the learning rate is tuned to—or very near—the optimal value.

\begin{table}[htbp]
  \centering
  \caption{$\mu$P formulation used in this work (adapted from \citep{yang+2021}). Initialization variance and learning rate scaling rules for SGD and AdamW under $\mu$P.}
  \label{tab:mup-scaling-rules}
  \begin{tabular}{lccc}
    \toprule
    & Input weights \& biases & Output weights & Hidden weights \\
    \midrule
    Init. Var. & $1/\mathrm{fan\_in}$ & 1 ($1/\mathrm{fan\_in}$) & $1/\mathrm{fan\_in}$ \\
    Multiplier & 1 & $1/\mathrm{fan\_in}$ (1) & 1 \\
    SGD LR & $\mathrm{fan\_out}$ (1) & $\mathrm{fan\_in}$ (1) & 1 \\
    Adam LR & 1 & 1 & $1/\mathrm{fan\_in}$ (1)\\
    \bottomrule
  \end{tabular}
\end{table}

\begin{figure}[htbp]
  \centering
  \begin{subfigure}[t]{0.48\textwidth}
    \centering
    \includegraphics[width=\linewidth]{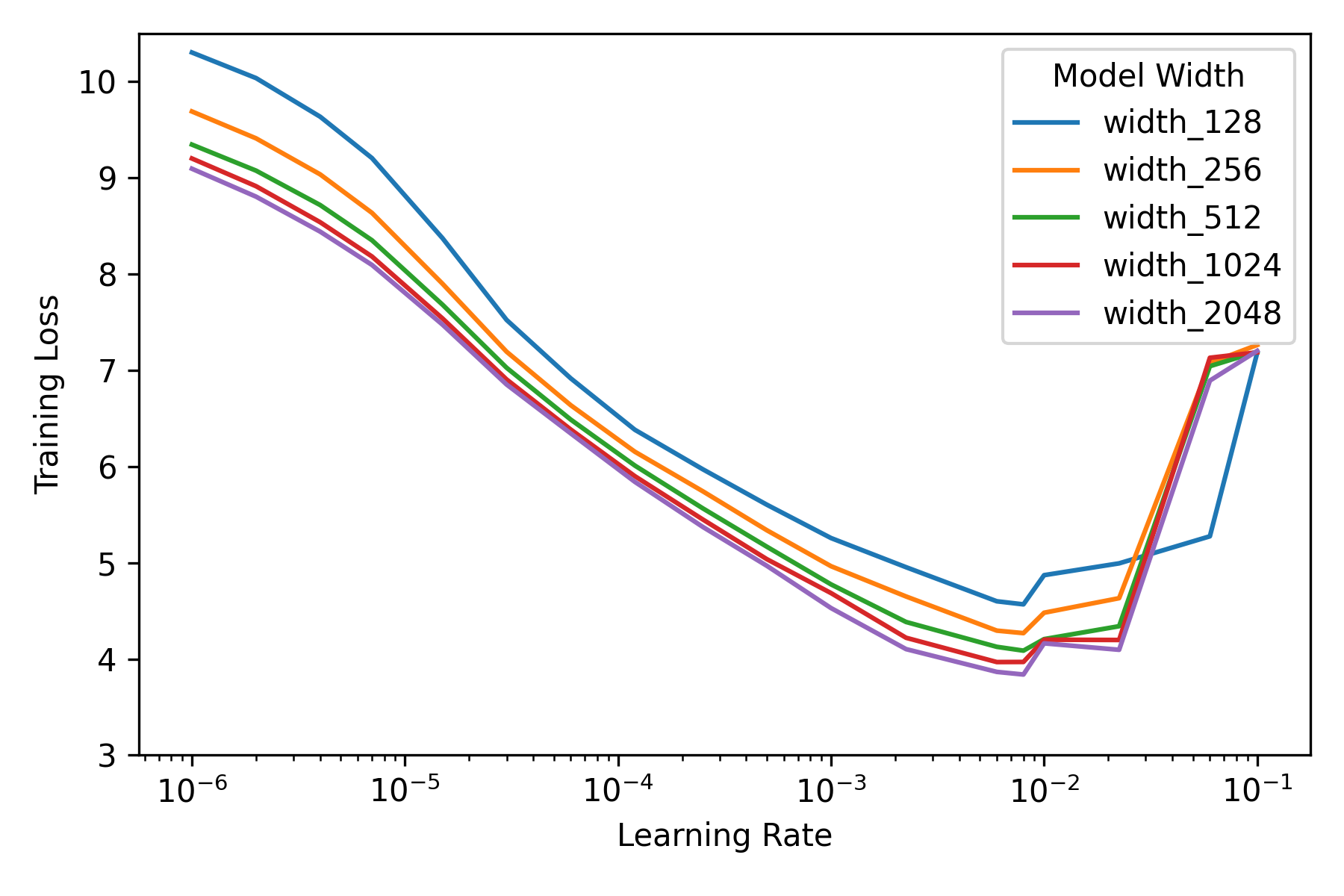}
    \caption{Lion}
    \label{fig:fig1}
  \end{subfigure}
  \hfill
  \begin{subfigure}[t]{0.48\textwidth}
    \centering
    \includegraphics[width=\linewidth]{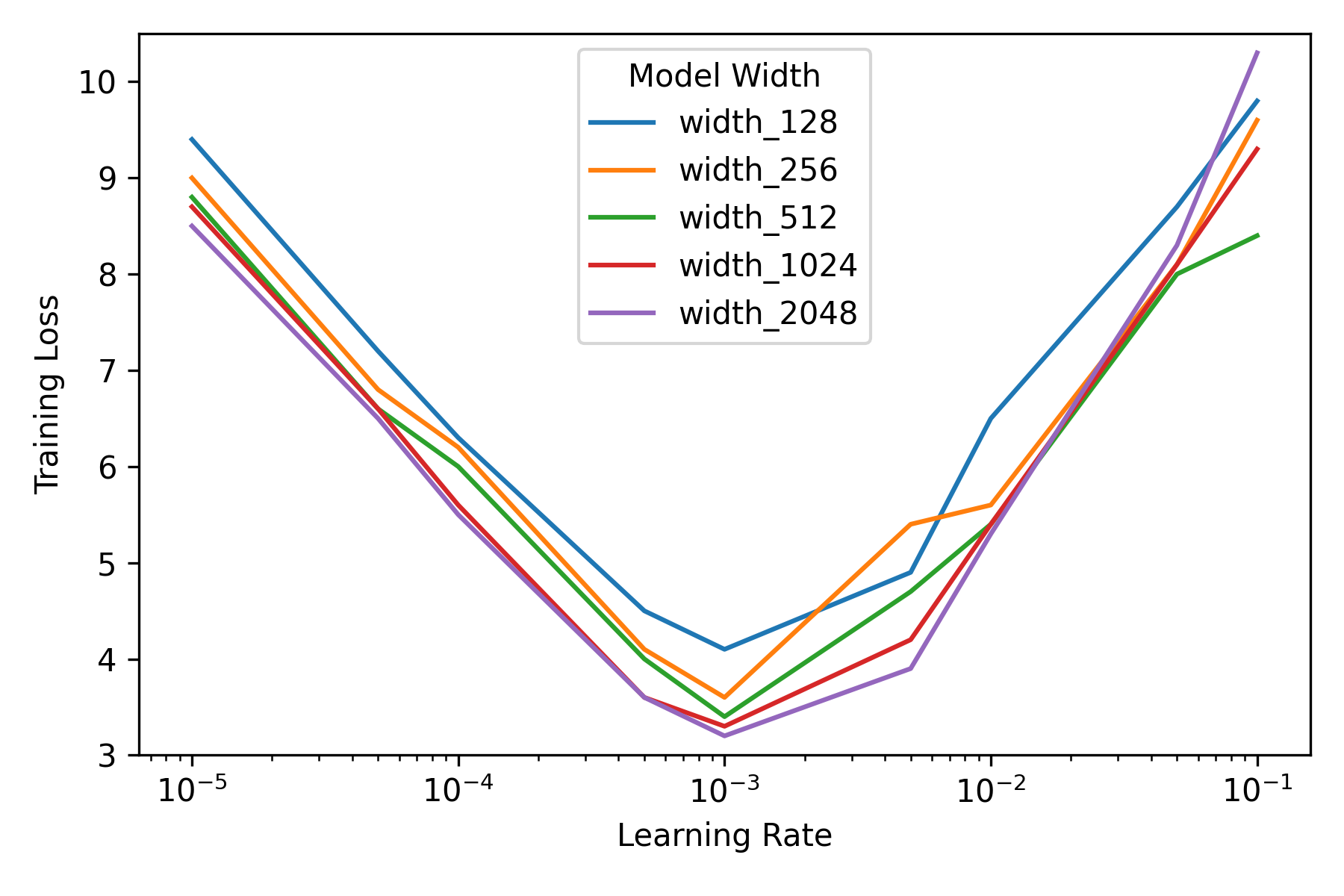}
    \caption{Sophia}
    \label{fig:fig2}
  \end{subfigure}
    \caption{$\mu$Transfer validation for Lion (left) and Sophia (right). Evidence of hyperparameter transferability is indicated by alignment of the optimal learning rate, i.e., minima in train loss curves occurring at similar learning rate values. \textit{Note:} Learning rate ranges are not directly comparable due to differing base model width, which affect the effective learning rate scale. This does not affect the validity of the $\mu$Transfer findings.}
    \label{fig:appendix-mup-validation}
\end{figure}

\subsection{Hyperparameters}
\label{appendix:hyperparameters}

\paragraph{Hyperparameter selection strategy}
\label{appendix:hyperparameter-selection}
Our choice to specifically tune learning rate and output multiplier was guided by an extensive hyperparameter sensitivity study performed in separate, ongoing research. While a detailed presentation of these results is beyond the scope of this paper, we briefly summarize our rationale here.

We systematically investigated the influence and interdependencies of various hyperparameters, including learning rate, batch size, initialization variance, weight decay, data shuffling, learning rate decay, warm-up duration, output multiplier, and beta parameters. Among these, the learning rate consistently exhibited the most substantial impact on model performance by a wide margin. The output multiplier showed a comparatively moderate influence but was notably correlated with optimal learning rate values.

Another factor limiting the scope of tuned hyperparameters is the fact that not all hyperparameters are $\mu$Transferable. This particularly applies to regularization-related hyperparameters such as weight decay \citep{yang+2021}.

Initialization variance was another critical parameter; however, prior research consistently identifies an optimal initialization variance value of approximately 0.073 for our tested models of the size considered here (a value also chosen by \citep{dey2023btlm} for a 3B parameter model), thus eliminating the need for further exploration within this study.

For the embedding multiplier, we adopted the recommendation of \citet{takase2025spike} to use the square root of the hidden dimension for enhanced training stability. Specifically, we set this multiplier based on the hidden dimension of our target model ($sqrt(2560) \approx 50.6$), achieving excellent training stability and validating this approach in our setting.

\paragraph{Optimizer-specific hyperparameter tuning}
Due to a fixed compute budget, we selectively tuned hyperparameters for each optimizer based on their sensitivity and computational cost. Specifically:

\begin{table}[ht]
\centering
\caption{Hyperparameters tuned per optimizer.}
\label{tab:optimizer_hparams}
\begin{tabular}{lccc}
\toprule
Hyperparameter & AdamW & Lion & Sophia \\
\midrule
Learning Rate & Yes & Yes & Yes \\
Output Multiplier & Yes & Yes & Yes \\
$\rho$ (Second-order effect) & No & No & Yes \\
\bottomrule
\end{tabular}
\end{table}

Sophia includes an additional parameter $\rho$, controlling the intensity of second-order approximation updates. Due to resource constraints, we slightly reduced the granularity of Sophia's learning-rate search to accommodate exploration of $\rho$.

\begin{figure}[t]
\centering
\includegraphics[width=0.8\linewidth]{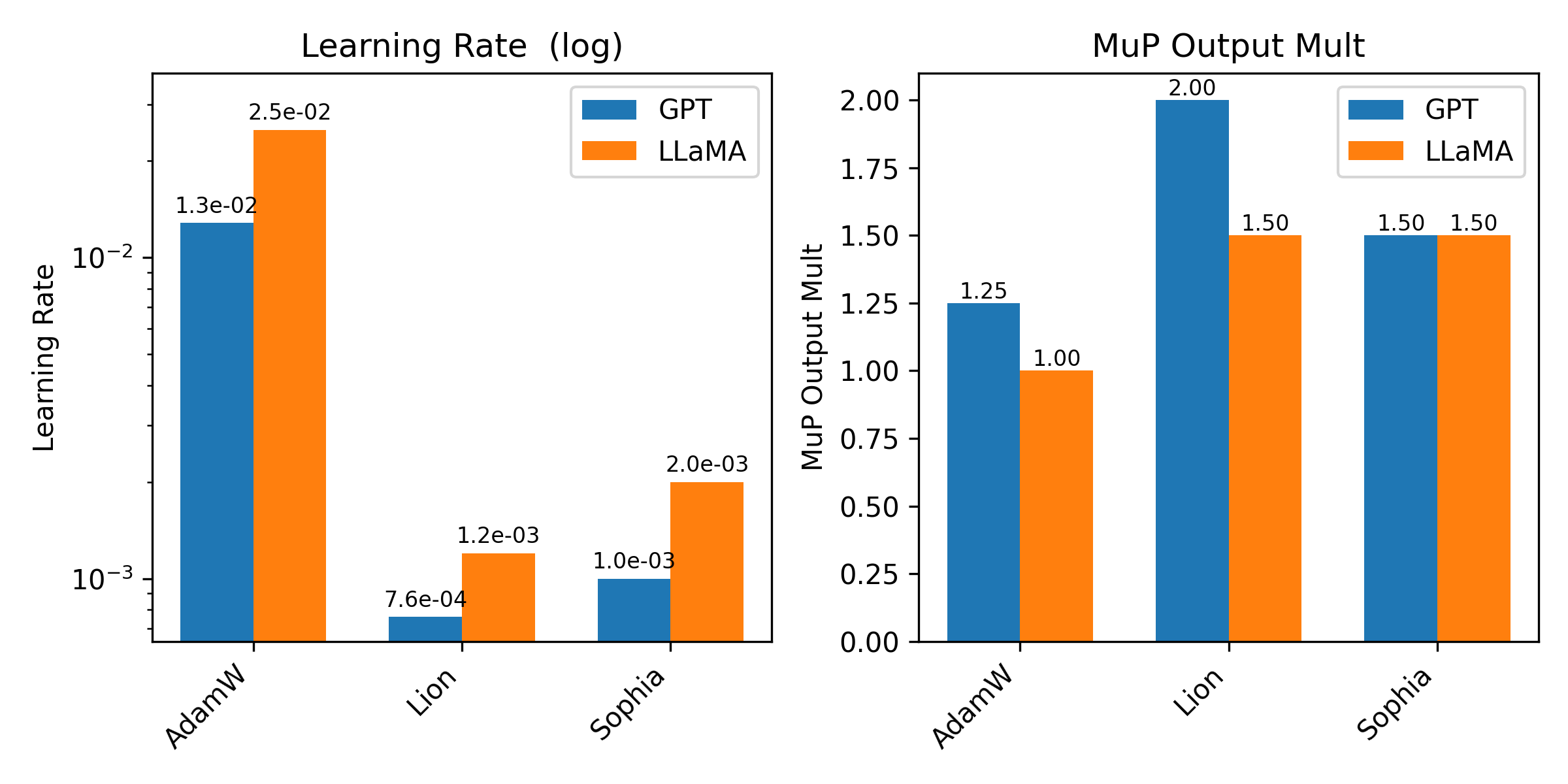}
\caption{Visualization of our tuned hyperparameters (learning rate and output multiplier) for each optimizer and architecture.}
\label{fig:appendix-hyperparameter-comparison}
\end{figure}

\paragraph{Final hyperparameters for large-scale training}
Hyperparameters for large-scale training were selected via systematic tuning on a smaller-scale proxy model (50M parameters), trained on an 1B-token subset of the SlimPajama dataset. We chose configurations based on minimal average training loss measured over the final 5\% of training steps. The final selected hyperparameters are summarized in Table~\ref{tab:hyperparams_summary}. An overview of the outcome of the tuned hyperparameters can be seen in Figure~\ref{fig:appendix-hyperparameter-comparison}.

\begin{table}[ht]
  \centering
  \caption{Hyperparameters for each optimizer. Global settings: data seed = 42, global seed = 42, grad clip = 1.}
  \label{tab:hyperparams_summary}
  \begin{tabularx}{\textwidth}{l *{6}{>{\centering\arraybackslash}X}}
    \toprule
    Hyperparameter       & AdamW-GPT & AdamW-LLaMA & Lion-GPT  & Lion-LLaMA & Sophia-GPT & Sophia-LLaMA \\
    \midrule
    Learning Rate*        & 0.0128    & 0.025       & 0.00076   & 0.0012    & 0.001      & 0.002 \\
    Output Multiplier*    & 1.5       & 1.0         & 2         & 1.5       & 1.5        & 1.5 \\
    Embedding Multiplier & 50.6      & 50.6        & 50.6      & 50.6      & 50.6       & 50.6 \\
    Init $\sigma$       & 0.073     & 0.073       & 0.073     & 0.073     & 0.073      & 0.073 \\
    $\beta_1$            & 0.9       & 0.9         & 0.9       & 0.9       & 0.96       & 0.96 \\
    $\beta_2$            & 0.95      & 0.95        & 0.99      & 0.99      & 0.99       & 0.99 \\
    $\epsilon$            & $10^{-8}$      & $10^{-8}$        & $10^{-8}$     & $10^{-8}$      & $10^{-15}$       & $10^{-15}$ \\
    Weight Decay         & 0.1       & 0.1         & 1.0      & 1.0       & 0.2        & 0.2 \\
    $\rho$*               &  -      &    -      &    -    &   -    & 0.3        & 1.0 \\
    \bottomrule
  \end{tabularx}
    \vspace{1ex}
\parbox{0.9\linewidth}{
  \footnotesize \textit{Note.} Due to time and budget constraints (and limitations of $\mu$P) we could not tune all hyperparameters. We focused on the most critical ones and set $\beta_1$, $\beta_2$, and weight decay to the authors' recommended values to avoid favoring any optimizer. Hyperparameters marked with '*' have been tuned.
}
\end{table}

\begin{figure}[t]
\begin{center}
\includegraphics[width=0.8\linewidth]{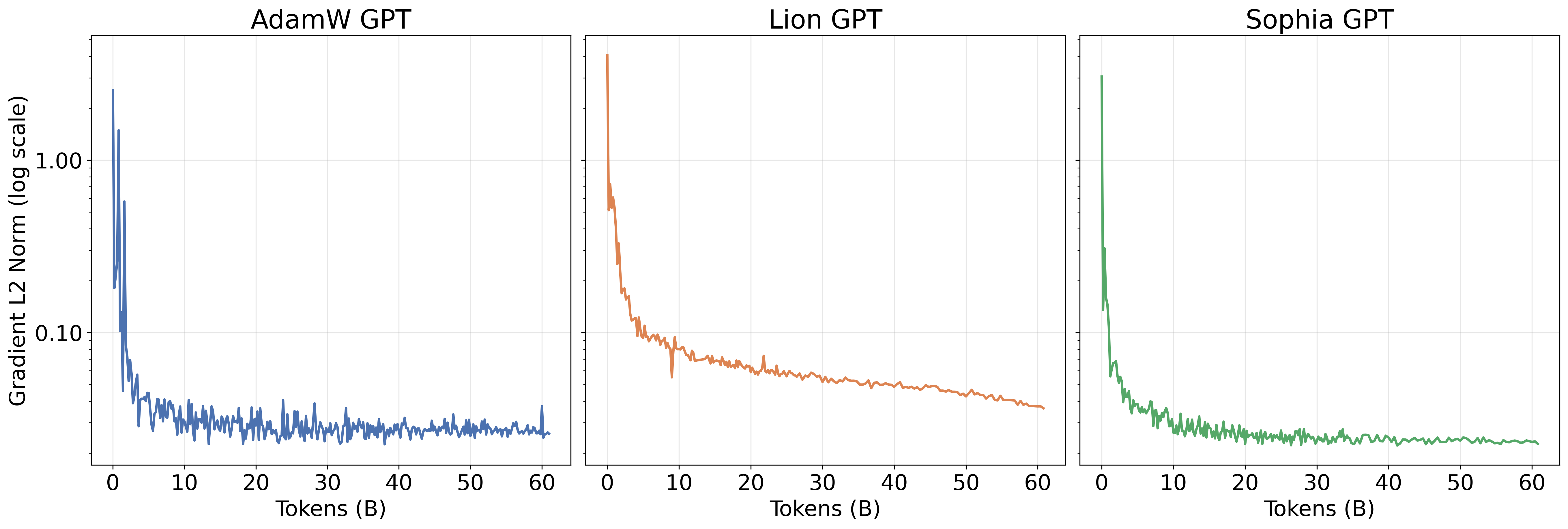}  
\end{center}
\caption{Average gradient L2‐norm (log scale) for GPT models trained with AdamW, Lion, and Sophia, plotted against the number of tokens processed.}
\label{fig:l2-gradient-norms}
\end{figure}

\end{document}